# Multiresolution Approximation of Polygonal Curves in Linear Complexity

Pierre-François Marteau, Gildas Ménier

*VALORIA, Université de Bretagne Sud, BP573, 56017 Vannes France*
*Phone. : (33) 2 01 72 99,  Fax : (33) 2 01 72 79*
{pierre-francois.marteau, gildas.menier}@univ-ubs.fr
http://www-valoria.univ-ubs.fr/

**Abstract**: We propose a new algorithm to the problem of polygonal curve approximation based on a multiresolution approach. This algorithm is suboptimal but still maintains some optimality between successive levels of resolution using dynamic programming. We show theoretically and experimentally that this algorithm has a linear complexity in time and space. We experimentally compare the outcomes of our algorithm to the optimal "full search" dynamic programming solution and finally to classical merge and split approaches. The experimental evaluations confirm the theoretical derivations and show that the proposed approach evaluated on *2D* coastal maps either show a lower time complexity or provide polygonal approximations closer to the input discrete curves.

***Keywords***: *Polygonal curves; Polygonal approximation; Dynamic programming; Data reduction; Compression; Multiresolution;*

## 1. Originality and Contribution

We provide a semi optimal efficient solution to the problem of approximating multidimensional discrete curves using a small number of linear segments. This solution when compared with previous existing approaches (Douglas-Peucker in *O(N)*, MergeL2 in *O(Nlog(N))*, Kolesnikov-Franti in $O(N^2/K)$, optimal dynamic programming solution in $O(K.N^2)$, where *N* is the size of the discrete input curve, and *K* the number of polygonal segments of the required approximation) either shows a lower time complexity or provides better polygonal approximations. We theoretically prove that the time complexity of our algorithm is (*O(N)*), as it is upper bounded by a linear function of *N* that is independent from the number of segments *K* of the final approximation . To our knowledge, the proposed algorithm is the best so far having a linear time complexity. Furthermore, this algorithm provides a set of nested polygonal approximations that realises a multiresolution representation of the input curve allowing post-processing at various resolution levels. The applications that are more and more resource demanding such as computer vision, shape analysis, data mining, etc, greatly benefit from low complexity algorithms able to simplify a complex curve into a simple shape characterized with few polygonal segments. We analyse in details the sensibility of the parameters that conditioned the behaviour of the proposed algorithm and provide experimentations on 2D geographic maps.

## 2. Introduction

Approximation of multi dimensional discrete curves has been widely studied essentially to speed up data processing required by resource demanding applications such as Computer Vision, Computer Graphics, Geographic Information Systems and Digital Cartography, Data Compression or Time Series Data Mining. For polygonal approximation of discrete curves, the problem can be informally stated as follows: given a digitized curve $X$ of $N \geq 2$ ordered samples, find $K$ dominant samples among them that define a sequence of connected segments which most closely approximate the original curve. This problem is known as the *min-ε* problem. Numerous algorithms have been proposed for more than thirty years to solve efficiently this optimisation problem. Most of them belong either to graph-theoretic, dynamic programming or to heuristic approaches.

Graph-theoretical applied to the *min-ε* problem produce a weighted directed acyclic graph (*DAG*) from the vertices (discrete points) of *X*, and then find the shortest path in this graph (Imai and Iri, 1986, 1988; Melkman and O'Rourke, 1988; Chan and Chin, 1996; Zhu, and Seneviratne, 1997; Chen and Daescu, 1998; Katsaggelos et al., 1998). For *min-ε* problem, finding a minimum path in the corresponding DAG can be solved in $O(N^2.\log(N))$ time (Chan and Chin, 1996) and in $O(N)$ space (Chen and Daescu, 1998).

Among *dynamic programming* solutions, Perez and Vidal (Perez and Vidal, 1994) were the first (to our knowledge) to propose an algorithm that exploits the sum of the squared Euclidean distance as the global error criterion. Their algorithm requires $O(K.N^2)$ time and $O(K.N)$ space. Some improvements have been proposed by Salotti (Salotti, 2001) to reduce the time complexity of this algorithm down to $O(N^2)$. Salotti's improvements consist of inserting a lower bound to limit the search space and employing the *A\** search algorithm instead of the dynamic programming one. Keeping with the ideas of Perez and Vidal, Kolesnikov et al. (Kolesnikov et al., 2004) introduced a 'bounding corridor', to limit the search space, and used an iterated dynamic programming within it to find an almost optimal solution. The time complexity is reduced to $O(W.N^2/K)$ where *W* is the size of the bounded corridor.

While dynamic programming and graph theoretic approaches target relatively optimal results, many algorithms try to relax optimality in order to lower the algorithmic complexity. Relying on the Diophantine definition of discrete straight line and its arithmetical characteristics, Debled-Rennesson and Reveillès (Debled-Rennesson and Reveillès, 2003 ) gave a linear method for segmentation of curves into exact discrete lines. Their idea is to extend a segment incrementally as much as possible so that the vertex that cannot be added to the segment becomes the lower extremity of the following segment. Similarly, Charbonier & al., (Charbonnier al., 2004) proposed an algorithm that splits a monitored signal into line segments—continuous or discontinuous—of various lengths and determines on-line when a new segment must be calculated: they used a cumulative sum (CUSUM) error criteria as the basis for their splitting heuristic. Pratt and Fink (Fink & Pratt, 2002) described a heuristic procedure for identifying major minima and maxima of a time series, and for their procedure proposed compression and time series information retrieval applications that could be used to extract line segments in linear time complexity.

This paper focuses on polygonal approximation of multidimensional curves using multiresolution for a given set of segment number for the crudest approximation level. Our main contribution is the development of an algorithm that, starting from the finest resolution level, finds min-epsilon polygonal approximation for more coarse resolution level using the approximating nodes obtained for the previous (more fine) resolution level. The number of line segments for the next, more coarse resolution level is reduced using a fixed factor $\rho$ in ]0;1[. Such multiresolution approximation can exploit any polygonal simplification methods between two successive levels of resolution, in particular heuristic algorithms (Douglas-Peucker, Merge-based algorithms, etc.). We address in this paper the use of algorithms based on constrained dynamic programming approach to ensure that the provided polygonal approximations are maintained close to optimal.

The first part of the paper describes the multiresolution algorithm we propose as an altenative solution to the *min-ε problem*. We show theoretically that the complexity of the algorithm is linear, both in time and space. The second part of the paper demonstrates experimentally the behaviour and the efficiency of our multi resolution procedure on a set of *2D* maps representing parts of the 'fractal' Brittany coast line (Mandelbrot, 1967). Finally, following previous works (Perrez and Vidal, 1994, Kolesnikov and al. 2004, Keogh and Pazzani, 2000) we present in the appendix the way we have specifically addressed the question of how to manage the polygonal approximation of curves using dynamic programming solutions for a single resolution level.

# 3. Multi resolution simplification of Multivariate Times series using polygonal curves approximation

As briefly explained in the introduction, several authors have already proposed to approximate polygonal curves using dynamic programming solutions for single resolution level.
Based on these earlier works, we present hereinafter a multiresolution algorithm that uses iteratively a constrained dynamic programming algorithm to find efficiently and sequentially polygonal approximations with minimal errors between each successive resolution levels.

### 3.1 Parameters and notation:

- *X(m)*: a discrete time series
- *p*: the dimension of the space that embeds *X*; $\forall m, X(m) \in R^p$
- *N*: the number of samples or length of a multivariate time series
- *K*: the number of segments of a polygonal approximation
- $\rho$: the ratio *K/N* : $\rho \in ]0;1[$
- *band* : half size of the bounded corridor used to reduce the search space of the *PyCA* algorithm (see FIG. 12). *band* is a positive integer.
- $\alpha$ : a parameter used for formal and experimental evaluations of time and space complexities. $\alpha$ is related to *band* and $\rho$:

  $band = Round(\alpha.\frac{N}{K}) \leq \alpha.\frac{N}{K} = \frac{\alpha}{\rho}$ . We take in practice $\alpha$ in {1,2, …}.

- *Lb(i) = band–i*; Lower bound used to limit the search space of the *PyCA* algorithm
- $C_{inf}(j)$: Corridor lower bound for the $j^{th}$ segment
- $C_{sup}(j)$: Corridor upper bound for the $j^{th}$ segment
- $\mathcal{C}_N$: Complexity of the *PyCA* algorithm
- $\mathcal{C}_N^{MR}$ : Complexity of the multi resolution *PyCA* algorithm

### 3.2 A Multi Resolution approach to Polygonal Curve Approximation (*MR-PyCA algorithm*)

Basically, the idea behind the multi resolution approach to polygonal curve simplification is to successively approximate previous approximations obtained by using some given simplification algorithm, this process being initiated from an original discrete time series. Following (Kolesnikov & al. 2004), we take a sequence of polygonal curves { $X_0$, $X_1$, $X_2$,…, $X_r$} as a multiresolution (multiscale) approximation of a *N*-vertex input curve *X*, if the set of curves {$X_i$}satisfies the following conditions:

i) A polygonal curve $X_i$ is an approximation of the curve *X* for a given number of segments $K_i$ (*min-ε problem*) or error tolerance $\varepsilon_i$ (*min-# problem*), where *i* is a resolution level (i=0,1,2,…, *r*).

ii) The set of vertices of curve $X_i$ for resolution level *i* is a subset of vertices of curve $X_{i-1}$ for the previous (higher) resolution level (*i-1*). The lowest resolution level *r* is represented by the coarsest approximation of *X*. The highest resolution level *i=0* is represented by the most detailed approximation (namely the original curve $X_0=X$) with the largest number of segments $K_0 = N$. ($N=K_0 > K_1 > K_2 >…> K_r$) or smallest error tolerance $\varepsilon_0=0$ for some distance measure (e.g. $L_2$) ($\varepsilon_0<\varepsilon_1<\varepsilon_2<…<\varepsilon_r$).

Thus, an approximation curve $X_i$ is either obtained by inserting new points into the approximation curve $X_{r+1}$, or, conversely, $X_{i+1}$ is obtained by deleting points from the approximation curve $X_i$. These two approaches have led to the development of two very popular heuristic approaches: respectively *Split* and *Merge* methods. In the Split approach, an iterative mechanism splits the input curve into smaller and smaller segments until the maximum deviation is smaller than a given error tolerance ε (*min-# problem*), or the number of linear segments equals to the given $K_i$ (*min-ε problem*) for the current resolution level *i*. A famous split method is the Douglas-Peucker algorithm (Douglas and Peucker, 1973); this algorithm is known to have a *O(K.N)* complexity; it has been used for multiresolution approximation in (Le Buhan Jordan & al., 1998, Buttenfield, 2002).

In the Merge approach (Pikaz and Dinstein, 1995, Visvalingam and Whyatt 1993), the polygonal approximation is performed by using a cost function that determines sequential elimination of the vertices with the smallest cost value, while the two adjacent segments of the eliminated vertex are merged into one segment. The approximation curve $X_i$ is obtained by discarding vertices from the curve $X_{i-1}$ until the desired number of vertices $K_i$ (*min-ε problem*) is reached. This merge approach is known to have a *O(N.log(N))* complexity.

There are two sources for error increasing in multiresolution approximation in comparison with individual polygonal approximation:

    1. In multiresolution approximation, vertices for the next level of resolution should be selected among the vertices available at the current level of resolution. In individual polygonal approximation for the levels we do not have this constraint.

    2. Non-optimality of algorithm used for *min-$\varepsilon$* polygonal approximation.

In multiresolution approximation, we cannot reduce errors related to the first reason, but with better algorithm for *min-$\varepsilon$* problems between successive levels of resolution one can expect to approach near to optimal solutions. This observation leads to the basis of the *MR-PyCA* algorithm we proposed. *MR-PyCA* algorithm relates to the *Merge* approach: we initiate the simplification process from the finest resolution level and iterate to obtain the crudest one, while discarding some vertices during each iteration using a constrained based dynamic programming approach. We present in the appendix how we have specifically addressed this "one step" simplification procedure to ensure the paper is self content. Basically, the *PyCA* algorithm we use during each simplification iteration can be seen as a special case of the so-called "Reduced Search Dynamic Programming" (*RSDP*) algorithm proposed in (Kolesnikov and Fränti 2003), for which the reduced search is confined inside a fixed sized corridor. The main slight difference between *PyCA* and *RSDP* lies in the way cost functions are evaluated: processing time is reduced for *PyCA* algorithm by cost of increasing distortion. *PyCA* is also adapted so that the simplification procedure can be called iteratively from the original time series down to the coarsest approximation level.

The pseudo-code for the multiresolution algorithm *MR-PyCA* is given in *FIG.1*.

**** FIG.1 around here ****

The inputs of *MR-PyCA* algorithm are:
- *K*, the number of segments in the polygonal approximation,
- *band*, the corridor width that reduces the search space,
- $\rho \in ]0;1[$, the decimation factor, that determines the fixed ratio of segments between two successive resolution levels,
- the original multidimensional time series $X=X_0$.

As *K* is an input, the number of resolution levels *r* (the number of iterations) is calculated given *K* and $\rho$. Given *K* and $\rho$, *r* is chosen such that:

$$N.\rho^{r+1} < K \leq N.\rho^r$$

As potentially $K < N.\rho^r$, a residual iteration is required to simplify the $r^{th}$ approximation (corresponding to resolution level $r$) that has $N.\rho^r$ segments in order to get an approximation having exactly $K$ segments. This last iteration discards ($N.\rho^r - K$) segments.

The multiresolution is the sequence of nested approximations provided in output. By construction this algorithm maintains partial optimality between two successive resolution levels, since a constrained dynamic algorithm (cf. Appendix) is used to search inside a fixed size 'corridor' for which segment extremities should be discarded and which should be kept. The approximation corresponding to the last resolution level is the *K-segments* polygonal approximation provided by *MR-PyCA*.

### 3.3 Complexity of MultiResolution *MR-PyCA*

For all *N*, *K<N* and $\rho$ in $]0;1[$ there exists a natural number *r* such that $N.\rho^{r+1} < K \leq N.\rho^r$, and defining $K_j = \rho^j.N$, then, using the *PyCA* algorithm while setting $band = round(\alpha.\frac{N}{K_1}) \leq \alpha.\frac{N}{K_1} = \frac{\alpha}{\rho}$, where $\alpha>1$ is a constant, we obtain from an original curve (*X*) of size *N* a polygonal curve approximation (*X₁*) having $K_1 = \rho.N$ segments with time complexity:

$$C_{N,1}^{MR} = 2.band^2.K_1 \leq \frac{2.\alpha^2.N^2}{K_1} = \frac{2.\alpha^2.N}{\rho}$$

If we consider as a second step the simplification of the $X_1$ curve still using the PyCA algorithm while setting $band = round(\alpha.\frac{K_1}{K_2}) \leq \alpha.\frac{K_1}{K_2} = \frac{\alpha}{\rho}$ then we get a polygonal curve approximation $X_2$ having $K_2 = \rho^2.N$ segments from the $X_1$ curve with time complexity:

$$C_{N,2}^{MR} = 2.band^2.K_2 \leq \frac{2.\alpha^2.N}{K_2} = 2.\alpha^2.N$$

Iterating the process with $band = round(\alpha.\frac{K_j}{K_{j+1}}) \leq \alpha.\frac{K_j}{K_{j+1}} = \frac{\alpha}{\rho}$ remaining constant, we get successively:

$$C_{N,3}^{MR} = 2.band^2.K_3 \leq \frac{2.\alpha^2.\rho}{\rho^2} = 2.\alpha^2.N$$

$$C_{N,4}^{MR} = 2.band^2 K_4 \leq \frac{2.\alpha^2.\rho^4}{\rho^2}.N = 2.\alpha^2.\rho^2.N$$

...

By induction, it is easy to show that for all $j$ in $\{1,..,r\}$ we have :

$$C_{N,j}^{MR} = 2.band^2 K_j \leq \frac{2.\alpha^2.\rho^j}{\rho^2} = 2.\alpha^2.\rho^{j-2}.N$$

For the final iteration required to ensure that the last approximation has exactly $K$ segments we use the PyCA algorithm setting $band = round(\alpha.\frac{K_r}{K}) \leq \alpha.\frac{K_r}{K}$.

Since $N.\rho^{r+1} < K$, we have: $\rho.(\rho^r.N) = \rho.K_r < K$ and then $\frac{K_r}{K} < \frac{1}{\rho}$. The time complexity for the last iteration is:

$$C^{MR}_{N,r+1} = 2.band^2 K \leq \frac{2.\alpha^2.K^2_r.K}{K^2} \leq 2.\frac{\alpha^2.\rho^r.N}{\rho} = 2.\alpha^2.\rho^{r-1}.N$$

Finally, from the original time series of size *N*, we get after *r* iterations of the previous process a polygonal approximation having *K* segments with time complexity:

$$C^{MR}_N = \sum_{j=1}^{r} C^{MR}_{N,j} \leq 2.\alpha^2.N.(1+\rho+\rho^2+...+\rho^{r-1}) = \frac{2.\alpha^2.N}{\rho}.\frac{1-\rho^r}{1-\rho} \quad (1)$$

as $\rho \in ]0;1[$, we get the following upper bound that shows that the time complexity of *MR-PyCA* when producing a *K* segments polygonal approximation from the original time series *X*:

$$C^{MR}_N \leq \frac{2.\alpha^2.N}{\rho.(1-\rho)} \quad (2)$$

We note that this upper bound is independent from *K*, showing that the time complexity of *MR-PyCA* is *O(N)*. This lower bound is furthermore minimized for $\rho = 1/2$.

The size of the search space required by the *PyCA* algorithm used during the $j^{th}$ iteration of *MR-PyCA* is included into a $K_j.(2.band)$ matrix. For $band \leq \frac{\alpha}{\rho}$ and $K_j = \rho^j.N$, the size of this matrix is $2.\alpha.\rho^{j-1}N \leq 2.\alpha.N$. For the first iteration (*j=1*), the space requirement for the matrix encoding is maximized upper bounded by $2.\alpha.N$. So, the search space required at any resolution level can fit into a $2.\alpha.N$ matrix allocated for the first resolution level. The polygonal approximation provided by the $(j-1)^{th}$ level of resolution is also required to compute the $j^{th}$ resolution level : the space requirement is $\rho^{j-1}N$. If we want to keep the approximations at all resolution levels we need to allocate a memory space that is upper bounded by $\sum_{j=0}^{r+1}\rho^{j-1}N = N.\frac{1-\rho^{r+1}}{1-\rho} \leq \frac{1}{1-\rho}.N$. The overall space requirement is thus upper bounded by $\left(2.\alpha + \frac{1}{1-\rho}\right).N$. This shows that the space complexity required for *MR-PyCA* is *O(N)*.

## 4. Experimentations and discussion

To evaluate the quality of suboptimal algorithms, Rosin (1997) introduced a measure known as fidelity (*F*). It measures how good (or how bad) a given suboptimal approximation is in respect to the optimal approximation in terms of the approximation error:

$$F = 100.\frac{E_{min}}{E},$$

where $E_{min}$ is the approximation error of the optimal solution, and *E* is the approximation error of the given suboptimal solution. In practice, we will identify $E_{min}$ to the error (the Euclidian distance between the original time series and the polygonal approximation) obtained using the 'Full Search' dynamic programming (*FSDP*) solution, namely the algorithm of Perez and Vidal, or alternatively our *PyCA* implementation with the *band* parameter set to *N*. The evaluation consists essentially in measuring the fidelity and runtime elapsed time for various parameter values of the *MR-PyCA* algorithm.

We have tested the *MR-PyCA* algorithm on 2D coastal maps extracted from the National Geophysical Data Center (NGDC, 2006) dataset. We have used essentially two maps: the Morbihan Gulf coastal map and the Brittany coastal map (*FIG 2*).

*\*\*\*\* FIG.2 AROUND HERE \*\*\*\**

*FIG. 3, FIG. 4* and *FIG. 5* compare respectively on the Morbihan Gulf 2D map (with *N=1478*) the optimal approximation solution with *K=33* segments in black continuous lines with the Douglas Peucker solution (*FIG. 3*), le *Merge_L2* solution (*FIG. 4*) and the *MR-PyCA* solution (*FIG. 5)*. The *MR-PyCa* solution (with $\rho=.85$, and $\alpha=4$) is clearly the closest to the optimal solution (*F=92,7%*) while the *Douglas-Peucker* algorithm provides an approximation with a significantly lower quality (*F=53%*) and the *Merge_L2* algorithm that provides an in between approximation (*F=77%*).

*\*\*\*\* FIG.3, 4, 5 around here \*\*\*\**

All the evaluation tests presented hereinafter have been performed on the Brittany coastal map (*FIG. 2)*. This *2D* map contains *17476* points given in longitude and latitude coordinates.

*FIG.6* shows the evolution of the error rate evaluated as the euclidian distance between the original map and the approximation map given by the *MR-PyCA* (with $\alpha=4$ and $\rho=.7, .75, .8$ and $.85$), *Merge_L2*, *Douglas Peucker* and *FSDP* algorithms as the resolution level decreases, e.g. *K* decreases. As expected, this error is a decreasing function of *K*. The *MR-PyCA* error curve is the closest to the optimal solution provided by the *FSDP* solution.

*\*\*\*\* FIG. 6 around here \*\*\*\**

*FIG.7* shows the evolution of the *F* measure as a function of *K* evaluated for the crudest approximation map given by the *MR-PyCA* (with $\alpha=8$ and $\rho=.1, .2, .3, .4, .5, .6, .7, .8$ and $.9$), *Merge_L2* and *Douglas Peucker*. *MR-PyCA* has a better *F* for all $\rho$ values. Furthermore, the higher $\rho$, the better *F* in general.

*\*\*\*\* FIG. 7 around here \*\*\*\**

*FIG. 8* shows the sensibility of the quality measure *F* while varying parameter $\alpha$ that defines the size of the fixed 'corridor'. This experiment shows that for a value which is too small for $\alpha$ ($\alpha < 4$) the quality of the approximations is poor for all *K*. The '*plateau*' of the curves indicates that it is not worth increasing the 'corridor' size too much: for $\alpha > 8$ the quality curves saturate for all *K* and no significant improvement is expectable.

*\*\*\*\* FIG. 8 around here \*\*\*\**

*FIG.9* compares the experimental time complexity of the optimal solution (*FSDP*), of the full search dynamic programming solution (*FSDP*), of the multiresolution algorithm (*MR-PyCA*), of the Douglas-Peucker algorithm (*DP*) and of the *Merge_L2* algorithm. The time complexity is measured as 'user clock ticks' spent as *N* increases while *K* remains fixed on a pentium 4 processor running Linux. The scale used in *FIG. 9* is logarithmic, so that all curves are linear with different slopes. The figure shows that the Douglas-Peucker complexity curve has the same slope as the *MR-PyCA* complexity curve. As the Douglas-Peucker algorithm is known to be *O(N)*, these two algorithms have the same linear complexity even though MR-PyCA is more expensive since the *MR-PyCA* curve is above the *DP* curve. *DPFS* curve has a higher slope, and as such exhibit a polynomial complexity as expected (*FSDP* is $O(N^2)$). The *Merge_L2* complexity curve has a slope in between FSDP and Douglas-Peucker curve as expected (*Merge_L2* is known to be *O(N.log(N))*).

*\*\*\*\* FIG.9 around here \*\*\*\**

*\*\*\*\* FIG.10 around here \*\*\*\**

*FIG. 10* shows the variations of the elapsed time for the *MR-PyCA* algorithm while varying parameter α. Clearly the complexity is polynomial with $\alpha$ meaning that enlarging the size of the 'corridor' implies an important time cost increase. For $\alpha$<32, the elapsed time remains small, and given that for $\alpha$ > 16 (as shown in *FIG. 9*), we do not have real improvement on the F measure, an optimal value for α is located around *8*.

*\*\*\*\* FIG.11  around here \*\*\*\**

In *FIG. 11* we evaluate the time complexity of the *MR-PyCA* algorithm in function of $\rho$, for various values of *N*. The experimentations shows that for all N values, the curves exhibit a minimum between $\rho=.30$ to $\rho=.55$. This roughly confirms the theoretical expectation (see eq. 11). The observed fluctuations are due to the fact that for the last iteration we use a varying value of $\rho$ to reach the required end value for *K*.

## 5. Conclusion

To our knowledge the proposed multiresolution solution applied to the problem of simplifying a curve using polygonal approximations is original. It consists in iteratively applying a constrained dynamic programming search algorithm on successive approximations of a polygonal curve. We have shown both theoretically and practically that this algorithm has a linear time complexity (*O(N)*), whatever the chosen number of resolution levels. This algorithm does not provide a single approximation, but a family of nested approximations from the finest to the crudest approximating levels with increasing distance between the original curve and the successive approximations. This algorithm is suboptimal but maintains partial optimality between each resolution levels. It offers good approximating solutions when real time and storage space are issues, namely each time the optimal solution cannot be calculated due to the size of *N*. For all tests we have performed, the quality of the resulting approximation is largely better than the quality of well known heuristic approaches (the Douglas-Peucker splitting approach or the merge approach): the gain on the quality measure *F* varies from *30%* to *50 %* according to the choice of the tuning parameters. The lowest quality measure that we have obtained is above *79%* (for small $\rho$, $\alpha$ and *K* values) while the best ones reach 100% (for large $\rho$, $\alpha$ and *K* values). The experimental results give highlights for the configuration of the tuning parameters of the algorithm, i.e. $\rho$, $\alpha$ that could vary according to the task. Furthermore, the multiresolution aspect of the method allows managing simultaneously various resolution levels, a functionality that could be very useful in time series information retrieval tasks.

# APPENDIX: Polygonal Curve Approximation using constrained Dynamic Programming (*PyCA*)

## 1.1 Problem formulation and "full search" dynamic programming solution

We consider time series as a multivariate process $X(t) = [x_1(t), x_2(t),..., x_p(t)]$ where $X(t)$ is a time-stamped spatial vector in $R^p$. In practice, we will deal with a discrete sampled time series $X(m)$ where $m$ is the time-stamp index ($m \in \{1,...,N\}$). Adopting a data modelling approach to handle the adaptive approximation of the time series, we are basically trying to find an approximation $X_{\hat{\theta}}$ of $X(m)$ such as:

$$\hat{\theta} = \underset{\theta}{ArgMin}(E(X, X_\theta)) = \underset{\theta}{ArgMin}\left(\sqrt{\sum_m \|X(m) - X_\theta(m)\|^2}\right) \quad (3)$$

where $E$ is the *RMS* error or Euclidian distance between $X$ and the model $X_\theta$.

In the case of polygonal curve approximation, we select the family $\{X_\theta(m)\}_{m \in \{1,...,N\}}$ as the set of piecewise linear and continuous functions (successive segments have to be contiguous, so that the end of a segment is the beginning of the next one). Numerous methods have been proposed for the problem of approximating multidimensional curves using piecewise linear simplification and dynamic programming in $O(k_N.N^2)$ time complexity (Perez et al. 1994). Some efficient algorithms (Goodrich, 1994) (Agarwal, 2002) with complexity $O(Nlog(N))$ have been proposed for planar curves, but none for the general case in $R^d$. Here, we have constrained the search of the segments by imposing that the extremities of the piecewise linear segments are vertices of time series $X(t)$. Thus, $\theta$ is nothing but the set of discrete time location $\{m_i\}$ of the segments' endpoints. Since the end of a segment is the beginning of the following one, two successive segments share a common $m_i$ at their interface. The selection of the optimal set of parameters $\hat{\theta} = \{\hat{m}_i\}$ is performed using a dynamic programming algorithm (Bellman 1957, Perez and Vidal 1994) as follows:

Given a value for $\rho$ and the size of the trajectory window to sample $N = |\{X(m)\}_{n \in \{1,...,n\}}|$, the number $K = |\{m_i\}| - 1$ of piecewise linear segments is known.

Let us define $\theta(j)$ as the parameters of a piecewise approximation containing $j$ segments, and $\delta(j,i)$ as the minimal error between the best piecewise linear approximation containing $j$ segments and covering the discrete time window $\{1,..,i\}$:

$$\delta(j,i) = \underset{\theta(j)}{Min}\left\{\sum_{m=1}^{i} \left\|X_{\theta(j)}(m) - X(m)\right\|^2\right\} \quad (5)$$

According to the Bellman optimality principle (Bellman, 1957) Perez and Vidal (Perez-Vidal, 1997) decomposed $\delta(j,i)$ as follows:

$$\delta(j,i) = \underset{m \leq i}{Min}\{d(m,i) + \delta(j-1,m)\}$$

where $d(m,i) = \sum_{l=m}^{i}\left\|\tilde{X}_{m,i}(l) - X(l)\right\|^2$ and $\tilde{X}_{m,i}(l) = (X(i) - X(m)).\frac{l-m}{i-m} + X(m)$ (6)

is the linear segment between *X(i)* and *X(m)*.

The initialization of the recursion is obtained given that: $\delta(1,1) = 0$.

The end of the recursion gives the optimal piecewise linear approximation, e.g. the set of discrete time locations of the extremity of the linear segments:

$$\hat{\theta}(K) = \underset{\theta(K)}{ArgMin}\left\{\sqrt{\sum_{m=1}^{N}\left\|X_{\theta(K)} - X(m)\right\|^2}\right\} \qquad (7)$$

with the minimal error :

$$\delta(K,n) = \sqrt{\sum_{m=1}^{N}\left\|X_{\hat{\theta}(K)}(m) - X(m)\right\|^2}$$

It is shown in (Vidal and Perez, 1994) that the complexity of the previous algorithm that implements a "Full Search" (FS) is in $O(K.N^2)$, a complexity that prevents the use of such an algorithm for large *N*.

## 1.2 "Constrained search" dynamic programming solution : the *PyCA* algorithm

In the scope of dynamic search algorithm the only way to reduce the time complexity is to reduce the search space itself. Sakoe and Shiba (Sakoe & Shiba, 1978) have managed to reduce the complexity of the Dynamic time Warping algorithm down to *O(N)* while defining fixed constraints that define a 'corridor' inside the search space. In the same mind-set, Kolesnikov and Fränti (Kolesnikov & Fränti, 2003) have developed locally adaptive constraints defining a varying width 'corridor' inside the search space to solve the problem of approximating polygonal curves using dynamic programming approach. Their algorithm shows to have complexity $O(W^2.N^2/K)$, where W is the size of their corridor. Following Sakoe and Shiba works, we have developed here a dynamic programming solution that implements a fixed size 'corridor': the search space is thus reduced using two fixed constraints, as shown in *FIG 12*. The first constraint limits the search of the $j^{th}$ segment upper extremity $i$ around the mean value $j.N/K$ namely the limit of the $j^{th}$ segment as to be chosen inside the interval:

$$[c_{inf}(j); c_{sup}(j)[, \text{ where}:$$
$$c_{inf}(j) = Max\{1, j.N/K - band\}; \quad (8)$$
$$c_{sup}(j) = Min\{N, j.N/K + band\}$$

The second constraint limits the search of the of the $j^{th}$ segment lower extremity $m$ in the interval $[Max\{1, Lb(i)\}; i[$, where $Lb(i) = (i - band)$.

*\*\*\*\* FIG12 Around here\*\*\*\**

Thus, the first constraint defines search bounds for the upper extremity of the $j^{th}$ segment (the $i$ index) while the second constraint defines a search bound for the lower limit of the $j^{th}$ segment $m \in [\max\{1, i - band\}; i[$, where *band* is fixed by the user. The recursive equations for the *PyCA* algorithm are then:

$$\delta(j,i) = \underset{lb(i) \leq m \leq i}{Min} \{d(m,j) + \delta(j-1,m)\}$$
$$\text{with } i \in [c_{inf}(j); c_{sup}(j)[,$$
$$c_{inf}(j) = Max\{1, j.N/K - band\}; \quad (9)$$
$$c_{sup}(j) = Min\{N, j.N/K + band\}$$

and with : $d(m,i) = \sum_{l=TY(m)}^{TY(i)} \left\| \tilde{X}_{TY(m),TY(i)}(l) - X(l) \right\|^2$

where $\tilde{X}_{TY(m),TY(i)}(l) = (X(TY(i)) - X(TY(m))).\dfrac{l - TY(m)}{TY(i) - TY(m)} + X(TY(m))$  (10)

with $TY(j)$ giving the time stamps correspondence of the $j^{th}$ segment extremity of the nested approximation $Y$ of $X$ in the original time series $X$ such that: $Y(j) = X(T\tilde{X}(j))$. *TY* is required for the multiresolution algorithm *MR-PyCA* that iteratively merges segments from the previous polygonal approximation to provide the next approximation. Indeed, for a direct use of PyCA, *TY* should be set to the identity relation such that *TY(j)=j*.

The initialization of the recursion is still obtained observing that: $\delta(1,1) = 0$ and the end of the recursion gives the **suboptimal** (it is optimal on the constrained search space) piecewise linear approximation, e.g. the set of discrete time locations of the extremities of the linear segments:

$$\delta(K,i) = \underset{lb(i) \leq m \leq i}{Min} \{d(m,K) + \delta(K-1, j)\}$$
$$\text{where } i \in [Max\{1, N - band\}; N[ \quad (11)$$

The pseudo code of the *PyCa* algorithm is presented in FIG. 13.

*\*\*\*\* FIG 13. Around here \*\*\*\**

## 1.3 Complexity of *PyCA*

According to the previous notations, the time complexity for the *PyCA* algorithm evaluates to:

$$C_N = 2.band^2.K \qquad (12)$$

If we choose $band = round(\alpha \frac{N}{K}) \leq \alpha \frac{N}{K}$ where $\alpha$ is a constant then

$$\frac{2.\alpha^2.(N - K/2)^2}{K} \leq C_N \leq \frac{2.\alpha^2.N^2}{K}$$ showing the time complexity of *PyCA* algorithm is $O(N^2)$.

The size of the search space for the *PyCA* algorithm is included into a *K.(2.band)* matrix. For $band = round(\alpha \frac{N}{K})$, the size of this matrix is upper bounded by $2.\alpha.N$ showing the space complexity of *PyCA* is $O(N)$.

# FIGURES and FIGURE CAPTIONS

```
//MR-PyCA Algorithm
MR-PyCA(K, α, ρ, X)
// K: the number of segments in the polygonal approximation
// band: The size of the corridor that delimits the search space
// α: The parameter that specifies the size of the corridor
//    that delimits the search space (band = ρ.N/K)
// ρ: the ratio of segments between two successive polygonal
       approximations. ρ is a constant.
// X: the input time series

// Initialisation:
set N ( length(X);// The length of the input time series
set r;            // such that  N.ρ^{r+1} < K ≤ N.ρ^r
set X_0 ← X;      // The finest resolution level is set to the
                  // original time series
N ← length(X);    // The length of the input time series
E_r ← 0;          // The approximation error for the finest
                  //resolution
// set the time stamps for the finest resolution level (0).
FOR i ← 1 TO N DO
   TX_0(i) ← i;
ENDFOR

// compute iteratively the crudest approximations PyCA
FOR i:= 1 TO r DO
  [E_i, X_i, TX_i](PyCA(X_0,X_{i-1},TX_{i-1}, ρ, α); // PyCA is described in
                                                    // the APPENDIX
ENDFOR

// if necessary compute the last approximation that has exactly K
// segments.
IF NOT( K = N.ρ^r )
  Set  ρ_0 ← K/(N.ρ^r)
  [E_{r+1}, X_{r+1}, TX_{r+1}](PyCA(X_r,X_1,TX_1, ρ_0, band);
  RETURN [[E_0, X_0, TX_0][E_1,X_1,TX_1], [E_2,X_2,TX_2],…,[E_{r+1},X_{r+1},TX_{r+1}]]
ELSE
  RETURN [[E_0,X_0,TX_0], [E_1,X_1,TX_1],…,[E_r,X_r,TX_r]]
ENDIF
// E_i is the L2 norm between the original time series and the
//    polygonal approximation at resolution level i.
// X_i is the sequence of samples that determine the segments
//    end points of the approximation at resolution level i.
// TX_i is the sequence of time stamps associated to each
//    segment end point of the approximation at resolution level i.
```

Fig. 1 – Pseudo code for the *MR-PyCA* algorithm.

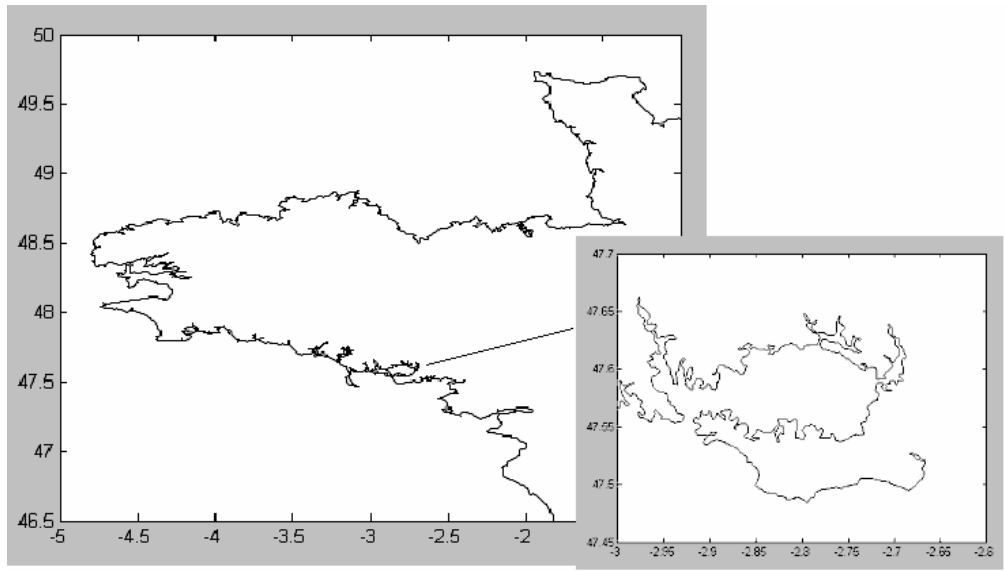

FIG. 2 –Morbihan Gulf and Britanny coastal 2D maps in longitude, latitude coordinates.

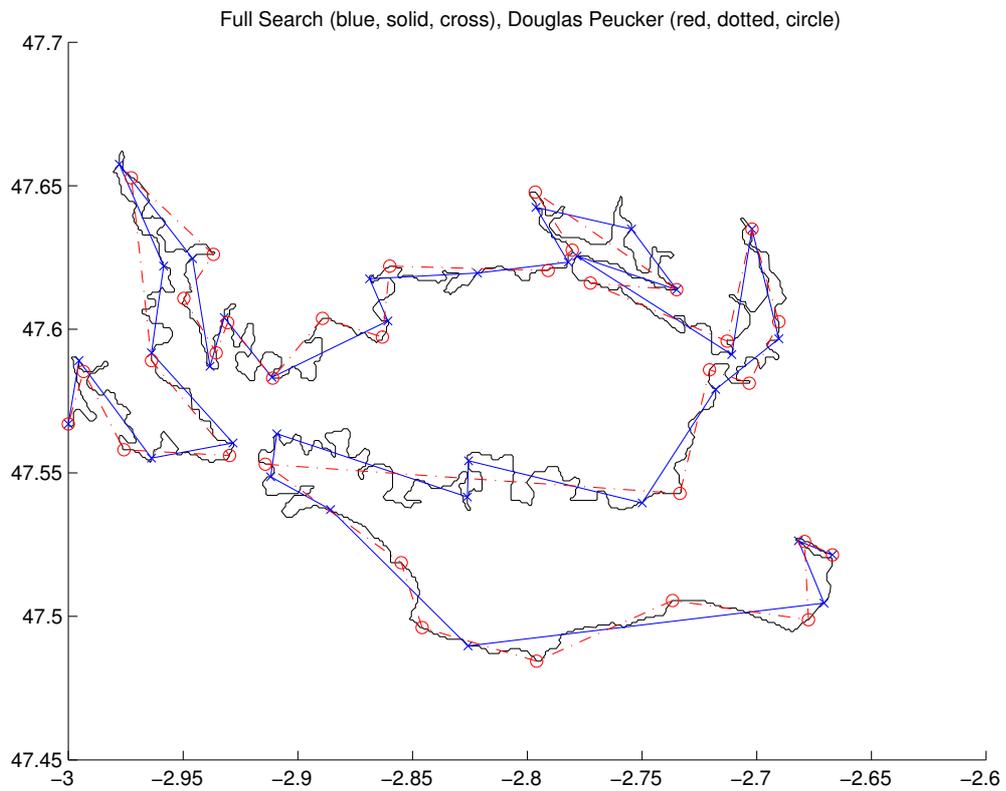

FIG. 3 –Morbihan Gulf 2D map in longitude, latitude coordinates: original map in black solid line, approximated map with *K=33* segments using the *FSDP* algorithm in blue dashed line, approximated map using *Douglas-Peucker* algorithm in red dotted line with circles.

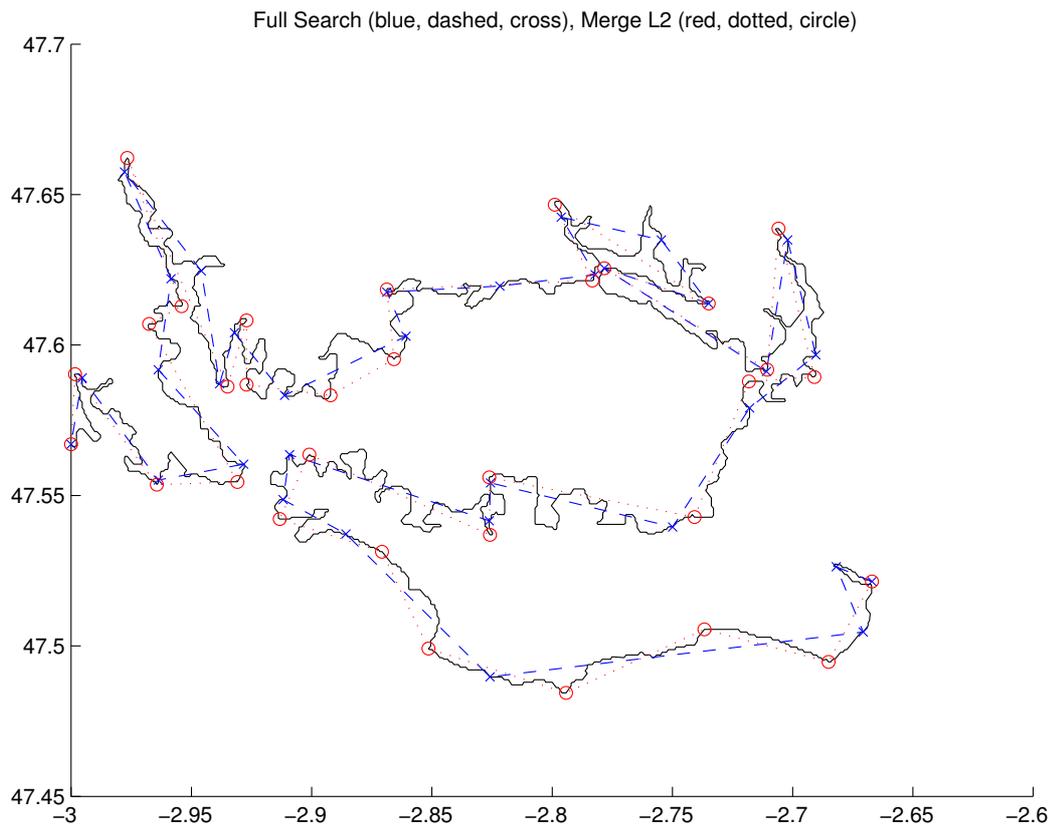

FIG. 4 –Morbihan Gulf 2D map in longitude, latitude coordinates: original map in black solid line, approximated map with *K=33* segments using the *FSDP* algorithm in blue dashed line, approximated map using *Merge_L2* algorithm in red dotted line with circles.

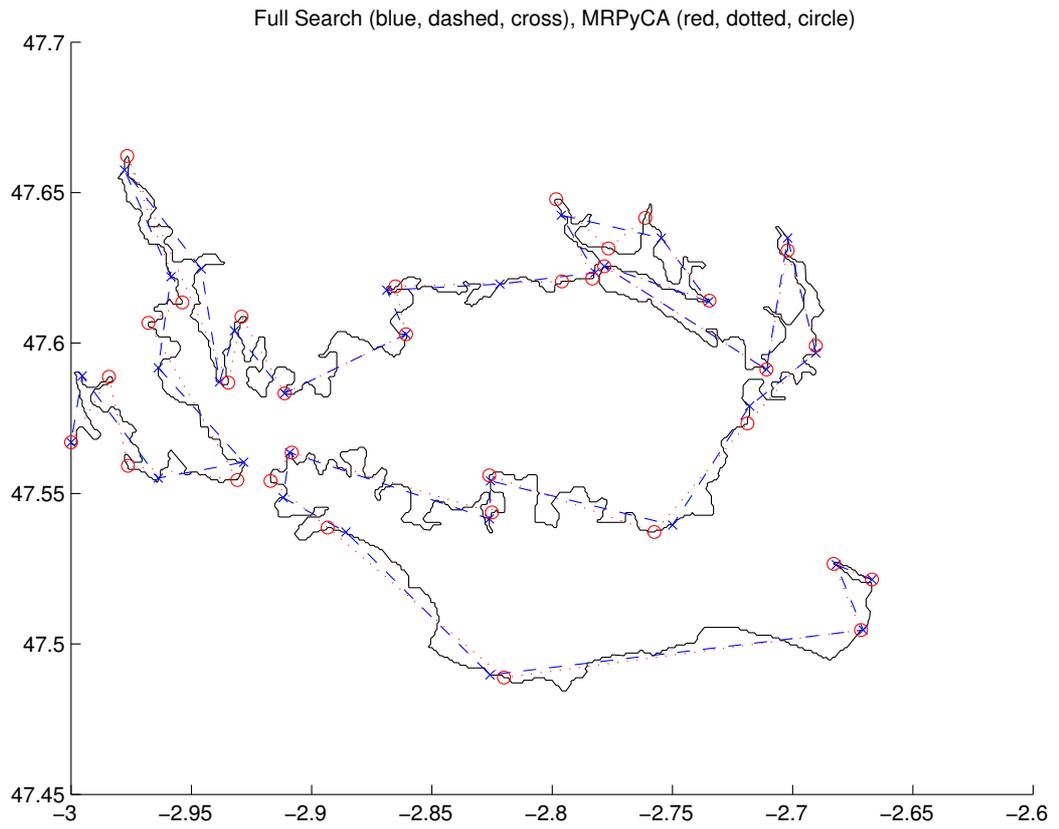

FIG. 5 –Morbihan Gulf 2D map in longitude, latitude coordinates: original map in black solid line, approximated map with *K=33* segments using the *FSDP* algorithm in blue dashed line, approximated map using *MR-PyCA ($\rho$=.8, $\alpha$=2)* algorithm in red dotted line with circles.

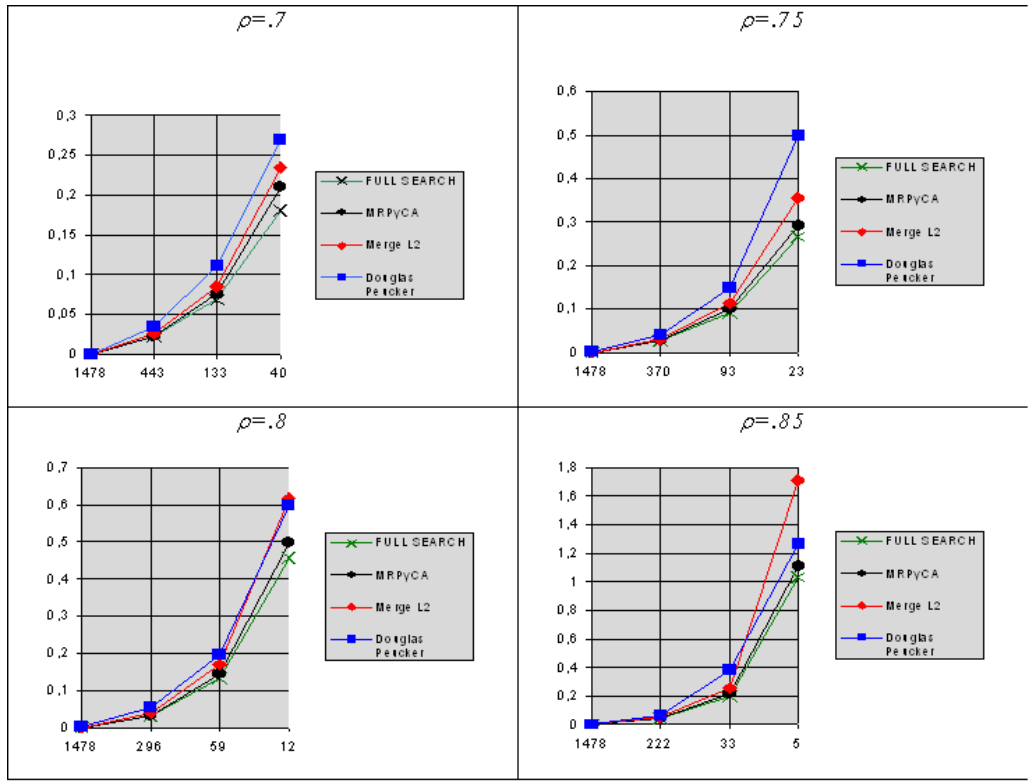

FIG. 6 – Euclidian distance between the original map and the approximated maps given by the *MR-PyCA (*with $\alpha=4$ and $\rho=0.7, 0.75, 0.8$ and $0.85$), *Merge_L2, Douglas Peucker* and *FSDP* algorithms as the resolution level decreases, e.g. *K* decreases.

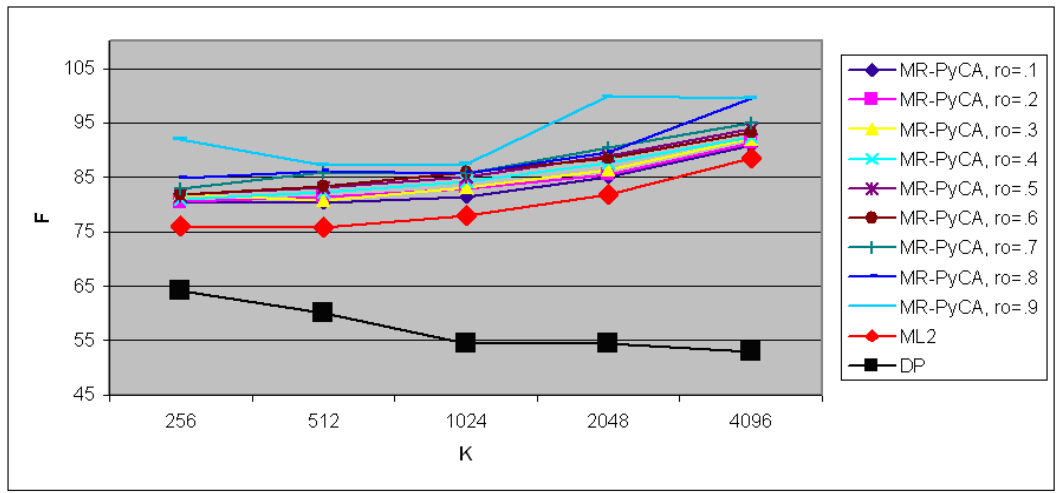

FIG. 7 – Fidelity measure *(F)* as a function of *K* evaluated for the crudest approximation map given by the *MR-PyCA (*with $\alpha=8$ and $\rho=.1, .2, .3, .4, .5, .6,.7, .8$ and $.9$), *Merge_L2* and *Douglas Peucker.*

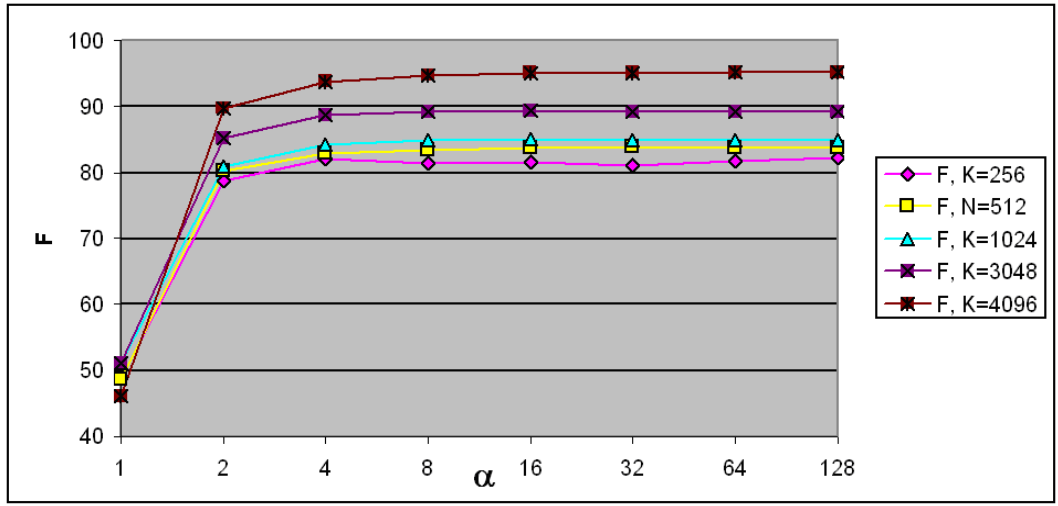

FIG. 8 – Fidelity measure (*F*) as a function of α and *K*.

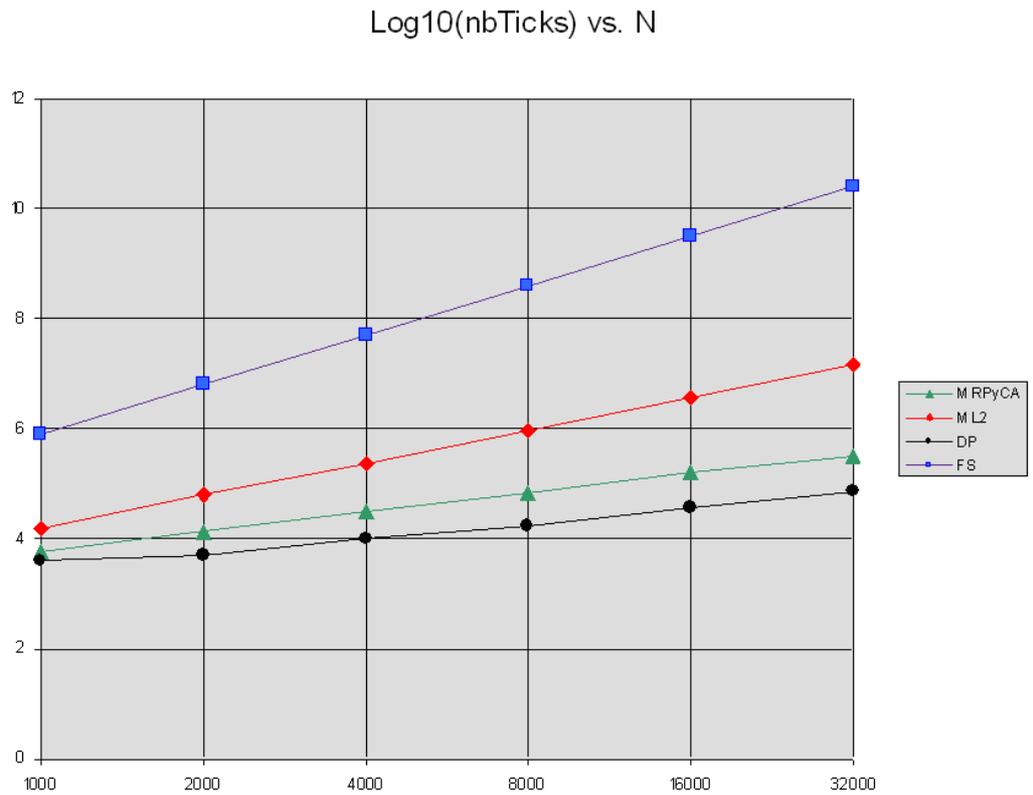

FIG. 9 –Comparison of experimental time complexity (expressed in user clock ticks) on a logarithmic scale for *Douglas-Peucker* algorithm (DP, circles, black), *MR-PyCA* algorithm (triangles, green), *Merge_L2* (rhombuses , red) and "*Full Search*" dynamic programming procedure (FS, squares, blue). Here *K = 10* for all methods, and ρ=.5, α=2 for *MR-PyCA*.

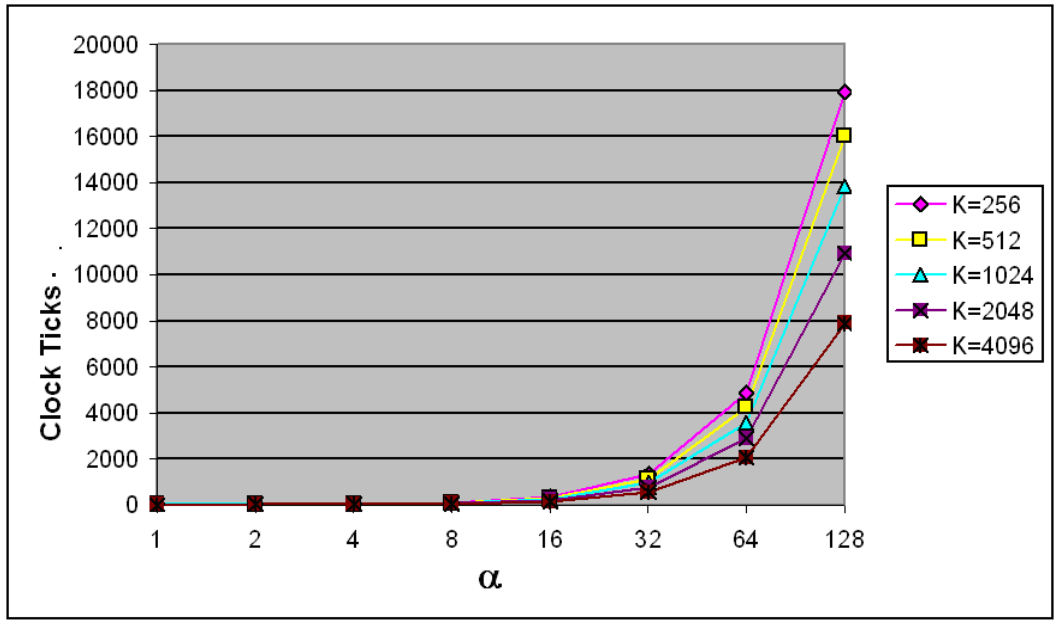

FIG. 10 – Time complexity (measured as a number of 'user clock ticks') as a function of $\alpha$ for a various number of segments $K$ in the crudest approximation of the multiresolution. The $\rho$ coefficient is kept constant ($\rho=1/2$).

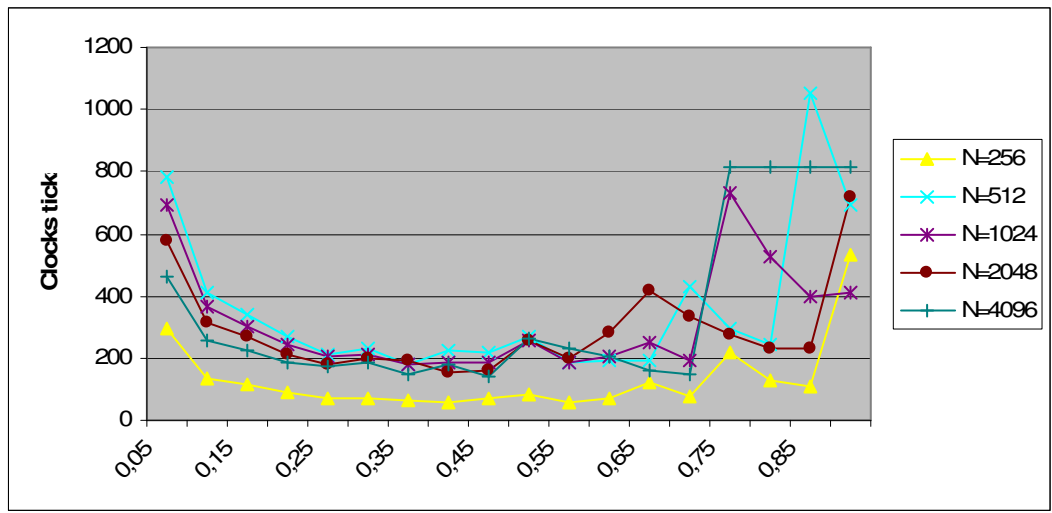

FIG. 11 – Time complexity (measured as a number of 'user clock ticks') as a function of $\rho$ for various number of segments $K$ in the crudest approximation of the multiresolution.

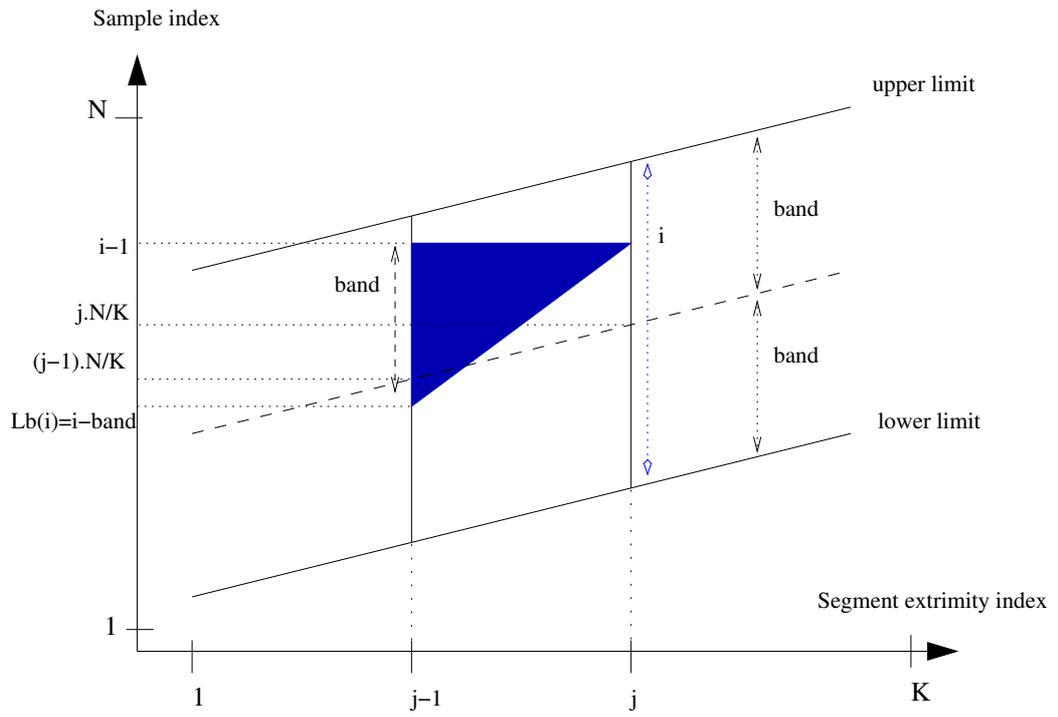

FIG. 12 – Limitation of the search space using upper and lower bounds for the search of the limits of the $j^{th}$ segment of the polygonal approximation.

```
PyCA(X, Y, TY, ρ, α)
// X: is the original time series
// Y: is a lower resolution of the X series
//    (potentially, Y equals to X ; if Y=X then TY(i)=i)
// TY: Time stamps correspondence for Y vertices in X time
series
// ρ: the ratio of segments between the input time series and
//    the polygonal approximation provided as ouput.
// α: The parameter that specifies the size of the corridor
//          that delimits the search space (band = α.N/K)

// Initialisation:
N ← Length(Y); // The length of the input time series
K ← ρ.N;   // the number of segments in the polygonal
           // approximation
band ← α.N/K;
δ(1,0) ← 0;
FOR i ← 2 TO N DO
   // Computation of the segmental approximation errors
   FOR m ← Lb(i) TO i-1 DO
```
$$d(m,i) \leftarrow \sum_{l=TY(m)}^{TY(i)} \left\| \tilde{X}_{TY(m),TY(i)}(l) - X(l) \right\|^2 \quad \text{// as defined in eq. (10)}$$
```
   ENDFOR //m
ENDFOR //i
// Compute iteratively the δ values
FOR j ← 1 TO k DO // loop on the segment index
   FOR i ← Cinf(j) TO Csup(j) DO //loop on the vertex index
      Emin ← ∞;
      FOR m ← Lb(i) TO i-1 DO
         E ← d(L(m),i) + δ(j-1,m);
         IF (E < Emin) DO
            Emin ← E;
            mmin ← m;
         ENDIF
         δ(j, i – Cinf(j)) ← Emin;
        Jstar(j, i – Cinf(j)) ← mmin;
      ENDFOR //m
   ENDFOR //i
ENDFOR //j
// Backtrack to extract the minimum path
j ← k; i ← N;
WHILE (j>0) DO
  is ← Jstar(j, i – Cinf(j));
Z(j) ← Y(is);
TZ(j) ← TY(is);
j ← j – 1; i ← is;
ENDWHILE
// Z is the sequence of samples that determine the k
//    segments end points of the approximation
// TZ is the sequence of time stamps associated to each
//    segment end point.
RETURN [δ(k,N), Z, TZ]
```

FIG. 13 –Pseudo code for the *PyCA* algorithm.